\documentclass[11pt,a4paper]{article}
\usepackage{authblk}
\usepackage[hyperref]{acl2020}

\usepackage{times}
\usepackage{latexsym}

\usepackage{graphicx}
\usepackage{float}
\usepackage{subfigure}
\usepackage{amsmath}
\usepackage{multirow}
\usepackage{booktabs}
\usepackage{algorithm}
\usepackage{algorithmic}
\usepackage{bm}

\newcommand{\tabincell}[2]{\begin{tabular}{@{}#1@{}}#2\end{tabular}}

\usepackage{microtype}

\aclfinalcopy 

\title{Self-attention Comparison Module for Boosting Performance on Retrieval-based Open-Domain Dialog Systems}

\makeatletter
\newcommand\email[2][]%
   {\newaffiltrue\let\AB@blk@and\AB@pand
      \if\relax#1\relax\def\AB@note{\AB@thenote}\else\def\AB@note{\relax}%
        \setcounter{Maxaffil}{0}\fi
      \begingroup
        \let\protect\@unexpandable@protect
        \def\thanks{\protect\thanks}\def\footnote{\protect\footnote}%
        \@temptokena=\expandafter{\AB@authors}%
        {\def\\{\protect\\\protect\Affilfont}\xdef\AB@temp{#2}}%
         \xdef\AB@authors{\the\@temptokena\AB@las\AB@au@str
         \protect\\[\affilsep]\protect\Affilfont\AB@temp}%
         \gdef\AB@las{}\gdef\AB@au@str{}%
        {\def\\{, \ignorespaces}\xdef\AB@temp{#2}}%
        \@temptokena=\expandafter{\AB@affillist}%
        \xdef\AB@affillist{\the\@temptokena \AB@affilsep
          \AB@affilnote{}\protect\Affilfont\AB@temp}%
      \endgroup
       \let\AB@affilsep\AB@affilsepx
}
\makeatother

\author[1]{\textbf{Tian Lan}}
\author[1]{\textbf{Xian-Ling Mao}}
\author[1]{\textbf{Zhipeng Zhao}}
\author[2]{\textbf{Wei Wei}}
\author[1]{\textbf{Heyan Huang}}
\affil[1]{Beijing Institute of Technology}
\email{\url{lantiangmftby@gmail.com},\url{oran\_official@outlook.com}\url{{maoxl,hhy63}@bit.edu.cn}}
\affil[2]{Huazhong University of Science and Technology}
\email{\url{Weiw@hust.edu.cn}}

\date{}

\begin{document}
\maketitle
\begin{abstract}
  Since the pre-trained language models are widely used, 
  retrieval-based open-domain dialog systems, 
  have attracted considerable attention from researchers recently. 
  Most of the previous works select a suitable response 
  only according to the matching degree 
  between the query and each individual candidate response. 
  Although good performance has been achieved,  
  these recent works ignore the comparison among the candidate responses,  
  which could provide rich information for selecting the most appropriate response. 
  Intuitively, better decisions could be made when the models can get access to the comparison information among all the candidate responses. 
  In order to leverage the comparison information among the candidate responses, 
  in this paper, 
  we propose a novel and plug-in \textbf{S}elf-attention \textbf{C}omparison \textbf{M}odule for retrieval-based open-domain dialog systems, 
  called SCM. 
  Extensive experiment results demonstrate that our proposed self-attention comparison module effectively boosts the performance of the existing retrieval-based open-domain dialog systems. 
  Besides, we have publicly released our source codes for future research\footnote{https://github.com/xxx/xxx}.
\end{abstract}

\section{Introduction}

Building a intelligent chatbot, or open-domain dialog system 
that can talk with human naturally is a very important task of the natural language process.
In terms of implementation, existing open-domain dialog systems can be divided into two categories: generative and retrieval-based.
In this paper, we only focus on the retrieval-based open-domain dialog systems.
By pre-training large scale language models, e.g. BERT \cite{Devlin2019BERTPO}, on vast corpus and subsequently fine-tuning these models on downstream corpus, 
recently, researchers have achieved state-of-the-art results in retrieval-based open-domain dialog systems 
\cite{DBLP:conf/iclr/HumeauSLW20,Tahami2020DistillingKF,Gu2020SpeakerAwareBF}.


The existing state-of-the-art retrieval-based open-domain dialog systems calculate 
the matching degrees between one given conversation context and multiple candidate responses,
and select the response that have the highest matching degree with the context.
Although great progress has been made, the existing works ignore the comparison information among the candidate responeses,
which could provide rich information for selecting the most appropriate response.
Intuitively, the model can make better decisions as long as it can get access to all of the candidate responses information, especially the comparison information among them.

In order to leverage the rich comparison information among the candidate responses,
in this paper, we propose a novel and plug-in \textbf{S}elf-attention \textbf{C}omparison \textbf{M}odule for the 
retrieval-based open-domain dialog systems, called SCM.
Specifically, we first separately concatenate the representation of the given conversation context and multiple candidate representations,
and use a transformer encoder \cite{Vaswani2017AttentionIA} to obtain the comparison information among them.
Then, the gated mechanism is introduced to obtain the final candidate response representations,
which blends the comparison information and each candidate response representation.
Finally, the matching degrees are calculated based on the conversation context representation and the final candidate response representations.

We modify two popular state-of-the-art retrieval-based open-domain dialog models by adding our proposed SCM module on them,
and conduct extensive experiments on three responses selection datasets.
Experiments results demonstrate that our proposed SCM module significantly boosts the performance on these two state-of-the-art models.

In this paper, our contributions are summarized as follows:

\begin{itemize}
  \item To our best knowledge, we are the first one to leverage the rich comparison information among candidate responses to boost the performance on retrieval-based open-domain dialog systems.
  \item We propose a novel and plug-in self-attention comparison module to effectively leverage the rich comparison information,
  which contains three important submodules.
  \item Extensive experiment results prove the effectiveness of our proposed SCM module, and our source codes have been publicly released for future research.
\end{itemize}

\section{Related Work}
\subsection{Retrieval-based Open-domain Dialog Systems}
Retrieval-based open-domain dialog systems, or response selection task, 
is a very important technique in the open-domain dialog systems.
Compared with the generative open-domain dialog system, 
retrieval-based open-domain dialog system is very easy to implement,
and could select more diverse and fluent responses from the pre-constructed corpus,
which is very popular in real application scenarios.

Over the past few years, 
retrieval-based open-domain dialog systems have been greatly developed 
and a large number of classical and powerful modeling methods have been proposed 
\cite{Wu2017SequentialMN,Zhang2018ModelingMC,Zhou2018MultiTurnRS,Tao2019MultiRepresentationFN,Gu2019InteractiveMN,Tao2019OneTO,Yuan2019MultihopSN}.
Especially with the help of the large scale pre-trained language model, e.g. BERT \cite{Devlin2019BERTPO},
the retrieval-based open-domain dialog systems achieve the state-of-the-art results.
So far, there are two kinds of approaches to leverage the pre-trained language models in retrieval-based open-domain dialog systems \cite{DBLP:conf/iclr/HumeauSLW20,Tahami2020DistillingKF}: 
(1) cross-encoder: cross-encoder \cite{Whang2019DomainAT,Gu2020SpeakerAwareBF} performs full (cross) self-attention over the concatenation of one given context and each candidate response.
(2) bi-encoder: bi-encoder \cite{DBLP:conf/iclr/HumeauSLW20,Tahami2020DistillingKF,Henderson2020ConveRTEA} first uses pre-trained language models to obtain the semantic representations of one given conversation context and multiple candidate responses separately.
Then the matching degree are obtained by calculating the dot production between one context representation and candidate representations.
\textbf{In this paper, we only focus on the bi-encoder models}, the reason are shown as follows.
For cross-encoders in the comparison settings, 
one given conversation context should be concatenated with all of the multiple candidate responses.
For example, 16 candidate responses (1 positive response and 15 negative responses) leads to 16 times increase in the size of training dataset,
which is very hard to train for pre-trained language model.
However, for bi-encoders, in-batch negatives approach \cite{Mazar2018TrainingMO,DBLP:conf/iclr/HumeauSLW20} allows for much faster and effective training in comparison settings.

In this paper, we choose two state-of-the-art bi-encoder models, bi-encoder and poly-encoder \cite{DBLP:conf/iclr/HumeauSLW20} as the backbone of our retrieval-based open-domain dialog systems,
and our proposed self-attention comparison module is added on them to boost the performance further.

\subsection{Transformer Architecture}
Transformer model \cite{Vaswani2017AttentionIA} has been proved to be a very effective architecture to 
process sequential data, for example, the natural language.
The transformer model consists of multiple blocks, and calculation flow of each block is shown as follows:
\begin{equation}
  \begin{split}
    & y = {\rm MultiHead}(x, x, x) \\
    & y = {\rm LayerNorm}(x, y) \\
    & z = {\rm FFN}(y) \\
    & z = {\rm LayerNorm}(y, z) \\
  \end{split}
\end{equation} where $x \in \mathcal{R}^{m\times d}$ is the input tensor 
($d$ is 768 BERT embedding size, $m$ is the length of the sequence). 
${\rm FFN}$ is a fully connected feed-forward network, 
and ${\rm LayerNorm}$ is the layer normalization mechanism \cite{Ba2016LayerN}.
The details of the multi-head self-attention ${\rm MultiHead}$ are:

\begin{small}
  \begin{equation}
    \begin{split}
      & {\rm Attention}(Q, K, V) = softmax(\frac{QK^T}{\sqrt{d_k}})V \\
      & h_i = {\rm Attention}(QW_i^Q,KW_i^K,VW_i^V) \\
      & {\rm MultiHead}(Q, K, V) = [h_1|h_2|...|h_k]W^O \\
    \end{split}
  \end{equation}
\end{small} 
where $k$ is the number of the heads, 
and $W^O, W^Q, W^k, W^V$ is the parameters of the transformer block.
$[\cdot|\cdot]$ denotes the concatenation operator.

For other natural language process task, the input tensor $x$ is usually the sequence that contains many token embeddings.
But in our proposed SCM model, the input tensor $x$ is a group of $m$ candidate response representations.

\subsection{Candidate Comparison}
To the best of our knowledge, 
the comparison information is first implemented in multi-choice reading comprehension task (MRC) \cite{Ran2019OptionCN,zhang2020dcmn+},
which requests to choose the right answer from a set of candidate options according to given passage and question.
However, there are two important difference between their methods and ours:
(1) In most of MRC corpus, there are very few candidate options according to the given passage and question \cite{zhang2020dcmn+},
for example four options in RACE corpus \cite{Lai2017RACELR}.
In the retrieval-based open-domain dialog systems, there are typically hundreds candidate responses,
and it is much more complex for retrieval-based open-domain dialog systems to leverage the comparison information;
(2) The model they proposed to leverage the comparison information is very complex to implement. 
In this paper, we use the popular transformer architecture \cite{Vaswani2017AttentionIA} to model the comparison information,
which is very easy and effective.

\begin{figure*}[h]     
  \center{\includegraphics[width=\textwidth, height=8cm] {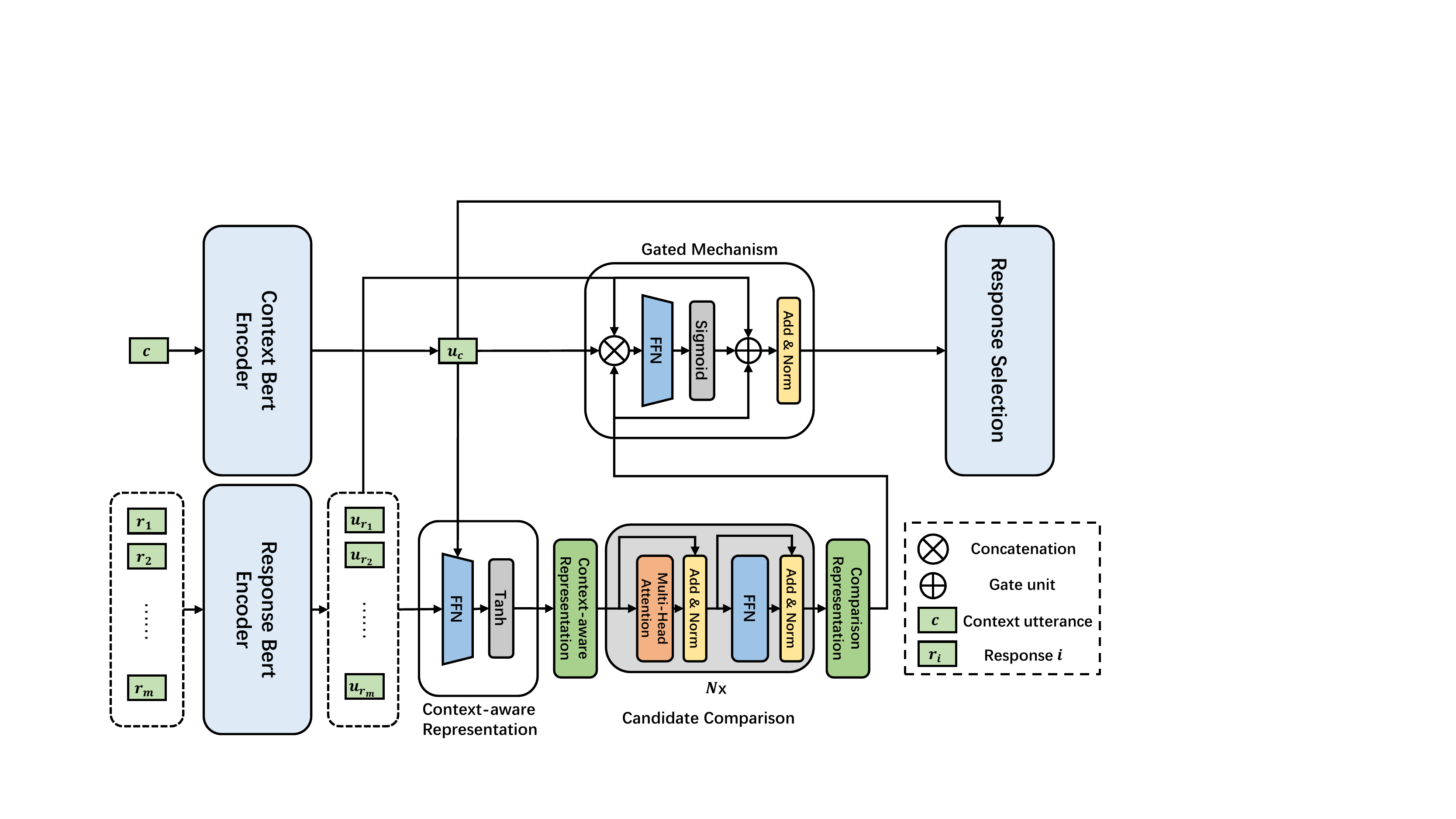}}        
  \caption{The overview architecture of our proposed retrieval-based open-domain dialog system that can leverage the comparison information among the candidate responses.
  It should be noted that the candidate comparison transformer model contains $N$ layers. 
  $\{r_i\}_{i=1}^m$ are $m$ candidate responses for the given conversation context $c$.}
  \label{img:1}
\end{figure*}

\section{Task Formulation}
Given one conversation context $c$ as the query and $m$ candidate responses $\{r_i\}_i^{m}$, 
the retrieval-based opend-domain dialog system should select one candidate $r_j$ as the final response,
which has the highest matching degree with $c$.

\section{Methods}

The overview of our proposed retrieval-based open-domain dialog system that leverages the comparison information is shown in Figure \ref{img:1}.
Our proposed model contains three modules: 
(1) Semantic representation module: it leverages the BERT model as the backbone to obtain the semantic representations of utterances (conversation context and candidate responses),
which is the same as the previous works \cite{Whang2019DomainAT,DBLP:conf/iclr/HumeauSLW20,Tahami2020DistillingKF};
(2) Self-attention comparison module: 
it contains three submodels to model rich comparison information among candidate response representations.: 
(a) context-aware representations; 
(b) candidate comparison;
(c) gated mechanism.
In Section \ref{sec:1}, we will describe these three submodules in details;
(3) Response selection: it use the context represenation and 
candidate response representations that contains rich comparison information 
to compute the matching degree,
and select the most appropriate candidate as the final response.

\subsection{Semantic Representation Module}
In this paper, we use two BERT \cite{Devlin2019BERTPO} models
to obtain the conversation context represenation and the candidate response representations separately.
Each BERT model has 12 layers, 12 attention heads, and a hidden size of 768.
The semantic representations can be obtained by:
\begin{equation}
    \begin{split}
      & u_c={\rm BERT_{context}}(c)\\
      & u_{r_i}={\rm BERT_{response}}(r_i), i\in \{1,2,...,m\}
    \end{split}
\end{equation} where $u_c, u_{r_i}\in \mathcal{R}^{768}$ is the semantic representations of conversation context and candidate responses $i$.
$m$ is the number of the candidate responses. 
${\rm BERT_{context}}$ and ${\rm BERT_{response}}$ are two BERT-base-chinese model \cite{Devlin2019BERTPO}.
It should be noted that these two BERT models parameters will be optimized separately during the training procedure.

\subsection{Self-attention Comparison Module (SCM)} \label{sec:1}
After obtaining the representations $u_c$ and $\{u_{r_i}\}_{i=1}^m$ by semantic represenation module,
we leverage the novel, easy, plug-in, and effective self-attention comparison module to model the comparison among the candidate response representations,
which has the following three submodules. 

\textbf{Context-aware Representations}:
First of all, before comparing the candidate responses, the conversation context $u_c$ must be considered.
Without considering the conversation context, the comparison information doesn't make sense.
In this paper, the context-aware representations are constructed by the following steps:
\begin{equation}
  \begin{split}
      & U_i=[u_c|u_{r_i}]\in \mathcal{R}^{768\times2} \\
      & H_i=tanh({\rm FFN}(U_i)), H_i\in \mathcal{R}^{768}
  \end{split}
\end{equation} 
where $\{H_i\}_{i=1}^m \in \mathcal{R}^{m\times768}$ is the context-aware representations, 
and $[\cdot|\cdot]$ denotes concatenation operator. 
$m$ is the number of the candidate responses.
${\rm FFN}$ is a fully connected network which squeezes the size of the representations from $768\times2$ to 768.

\textbf{Candidate Comparison}:
Then, we leverage the popular transformer model \cite{Vaswani2017AttentionIA} that has multiple layers
to process context-aware representations $\{H_i\}_{i=1}^m \in \mathcal{R}^{m\times768}$,
and obtain the comparison information among all the candidate responses.
The computation procedure is already shown in formula (1), 
and $\{H_i\}_{i=1}^m \in \mathcal{R}^{m\times768}$ is the input tensor $x$.
With the help of the multi-head self-attention mechanism in the transformer model,
each candidate response can receive the representations from other candidate responses,
and capture the difference among them effectively and elegantly.
After the processing by the self-attention comparison module,
The comparison information $\{O_i\}_{i=1}^m \in \mathcal{R}^{m\times768}$ is obtained.

\textbf{Gated Mechanism}:
Although $\{O_i\} \in \mathcal{R}^{m\times768}$ is already a good representations of candidate responses,
it may lose some important information that is in the original represenation of each candidate. 
In order to effectively combine the original representations and the comparison information, 
we employ the gated mechanism \cite{Hochreiter1997LongSM}:
\begin{equation}
  \begin{split}
      & o_i = [u_{r_i}|u_c|O_i], o_i \in \mathcal{R}^{768\times3} \\
      & g_i = sigmoid({\rm FFN}(o_i)), g_i \in \mathcal{R}^{768} \\
      & f_i = {\rm LayerNorm}(g_i * u_{r_i} + (1 - g_i) * O_i) \\
  \end{split}
\end{equation}
where ${\rm FFN}$ is the fully connected network that squeezes the size of $o_i$ from $768\times3$ to 768.
$g_i$ is a gate unit, which learns to save the important information of the original representations and comparison information.
$*$ is the element-wise product operator.
Finally, $\{f_i\}_{i=1}^m \in \mathcal{R}^{m\times768}$ are obtained, which contains both candidate response representations and the comparison information among them.

\subsection{Response Selection}
So far, we already have the context representations $u_c \in \mathcal{R}^{768}$ 
and the candidate representations $\{f_i\}_{i=1}^m \in \mathcal{R}^{m\times768}$.
Then, the matching degrees can be obtained by calculating the dot production 
$\{degree_i\}_{i=1}^m=\{f_i\}_{i=1}^m\cdot u_c \in \mathcal{R}^m$.
During training, we minimize the cross-entropy loss:
\begin{equation}
  \begin{split}
      & \{prob_i\}_{i=1}^m = softmax(\{degree_i\}_{i=1}^m) \\
      & \text{loss} = -\sum_{i=1}^m prob_i\log(p_i) \\
  \end{split}
\end{equation}
, where $m$ is the number of the candidate responses, 
and $p_i$ is the label of the candidate response $i$.
If $i$-th candidate response is the ground-truth response, $p_i=1$,
otherwise, $p_i=0$.
During inference, 
the candidate response $i$ that has the highest $degree_i$ will be sampled as the final response 
to the given conversation context $c$. 

\section{Experiemnts}
\subsection{Datasets}

We test our proposed model on two widely used multi-turn response selection datasets,
the E-Commerce Corpus \cite{Zhang2018ModelingMC} and Douban Corpus \cite{Wu2017SequentialMN},
and one open-domain dialog dataset Zh50w\footnote{\url{https://github.com/yangjianxin1/GPT2-chitchat}}.
Data statistics are in Table \ref{tab:1}.

\begin{table}[h]
  \begin{center}
    \resizebox{0.5\textwidth}{!}{
      \begin{tabular}{|c|c|c|c|c|c|c|}
      \hline
      \multirow{2}{*}{\textbf{Data statistics}}          & \multicolumn{2}{c|}{\textbf{Douban}} & \multicolumn{2}{c|}{\textbf{E-Commerce}} & \multicolumn{2}{c|}{\textbf{Zh50w}} \\ \cline{2-7} 
                                          & \textbf{Train}        & \textbf{Test}       & \textbf{Train}          & \textbf{Test}         & \textbf{Train}       & \textbf{Test}       \\ \hline
      \textbf{Sessions number}            & 0.5M                  & 667                 & 0.5M                    & 1K                    & 0.497M               & 3K                  \\ \hline
      \textbf{Avg turns per context}      & 7.69                  & 7.23                & 6.51                    & 6.64                  & 4.1                  & 3.96                \\ \hline
      \textbf{Avg words per utterance}    & 24.73                 & 26.59               & 11.37                   & 11.54                 & 10.25                & 11.95               \\ \hline
      \end{tabular}
    }
    \caption{Data statistics of Douban, E-Commerce, Zh50w Corpus.}
    \label{tab:1}
  \end{center}
\end{table}

\textbf{Douban Corpus} contains dyadic dialogs crawled from the Douban group\footnote{\url{https://www.douban.com/group}},
which is a popular social networking service in China.
It should be noted that each conversation context in Douban Corpus test dataset may have multiple ground-truths, 
and we ignore these cases in this paper, which squeezes the test dataset size from 1000 to 667. 

\textbf{E-Commerce Corpus} is collected from real world conversations 
between customers and customer service staff from Taobao\footnote{\url{https://www.taobao.com}},
the largest ecommerce platform in China.

\textbf{Zh50w Corpus} is a chinese open-domain dialog corpus between two persons. 
Compared with Douban Corpus and E-Commerce Corpus, the conversations in Zh50w Cropus are more casual.
It should be noted that the negative samples in test dataset of Zh50w Corpus 
are retrieved from Elasticsearch\footnote{\url{https://github.com/elastic/elasticsearch}} index set, which is the same as the Douban Corpus.

It should be noted that, for Douban Corpus and E-Commerce Corpus train datasets,
there are one positive and one negative sample for each conversation context.
But in this paper, we select the bi-encoder instead of the cross-encoder as our base model,
which treats responses in the same batch as the negative samples \cite{DBLP:conf/iclr/HumeauSLW20},
and doesn't need the negative samples in the original train dataset.
So the size of E-Commerce and Douban Corpus train datasets are squeezed from 1M to 0.5M.

\subsection{Evaluation Metrics}
Following the previous works 
\cite{Zhang2018ModelingMC,Zhou2018MultiTurnRS,Tao2019MultiRepresentationFN,Gu2019InteractiveMN,Tao2019OneTO,DBLP:conf/iclr/HumeauSLW20},
we employ recall at position k in $m=10$ candidates ($R_n@k$) and MRR (Mean Reciprocal Rank) as the evaluation metrics.
It should be noted that we also conduct the extended experiments on these datasets,
which is closer to the real scene settings.
In the extended experiments, each conversation context have more candidate responses in test dataset
($m$ is larger than 10, and $n$ is 50, 100, 150, 200, 250, 300).

\subsection{Parameters Settings}

All the models in this paper was implemented by PyTorch \cite{Paszke2017AutomaticDI}.
We use the transformers toolkit \cite{Wolf2019HuggingFacesTS} to construct the BERT model, 
and the BERT-base-chinese model parameter is used because the datasets are chinese.
The apex toolkit\footnote{\url{https://github.com/NVIDIA/apex}} is used to speed up.
More parameters can be found in Table \ref{tab:2}. 

\begin{table}[h]
  \begin{center}
      \resizebox{0.5\textwidth}{!}{
              \begin{tabular}{|l|c||l|c|}
                \hline
                \textbf{Param} & \textbf{Value} & \textbf{Param} & \textbf{Value} \\ \hline \hline
                BERT learning ratio    & 5e-5   & BERT embedding size     & 768   \\ \hline
                gradient clip  & 1.0            & epoch          & 5              \\ \hline
                batch size     & 16             & random seed    & 50             \\ \hline
                dropout ratio  & 0.1            & warmup step ratio & 0.1         \\ \hline
                ${\rm Poly}_m$ & 16             & max utterance length & 256 \\ \hline
                SCM encoder layer & 4 & SCM head number & 8  \\ \hline
                SCM dim ffd & 512 & SCM learning ratio & 5e-4 \\ \hline
              \end{tabular}
      }
      \caption{Parameters in this paper.}
      \label{tab:2}
  \end{center}
\end{table}

It should be noted that, for \textbf{bi-encoder} models,
the bigger batch size (bigger negative samples) is used, 
the better performance can be achieved.
In this paper, due to the memory size limitation of our devices,
we select the batch size 16.

\subsection{Models}
We select two state-of-the-art bi-encoder models \cite{DBLP:conf/iclr/HumeauSLW20}: 
\textbf{bi-encoder} and \textbf{poly-encoder} models, as the base models.
As shown in Figure \ref{img:1}, 
the \textbf{bi-encoder} and \textbf{poly-encoder} models only have the context bert encoder, 
response bert encoder, and the response selection module.
Then we add our proposed self-attention comparison module (SCM module) on these two base models,
which have three important submodules: (1) context-aware represenation; (2) candidate comparison; (3) gated mechanism,
denotes as \textbf{bi-encoder+SCM} and \textbf{poly-encoder+SCM}.

Besides, we also conduct the ablation study to analyze the contributions of two submodules in our proposed SCM module:
(1) context-aware represenation; (2) gated mechanism.
Specifically, we first remove each submodule in \textbf{poly-encoder+SCM} model,
which leads to two new models:
(1) \textbf{poly-encoder+SCM-\{context-aware\}};
(2) \textbf{poly-encoder+SCM-gated}.
Then their performances are measured on three datasets.
It should be noted that we don't remove the \textbf{candidate comparison} submodule,
the reasons are as follows:
(1) The \textbf{candidate comparison} is the core module in our proposed SCM module,
and it doesn't make sense to remove this submodule;
(2) The \textbf{candidate comparison} is between the other two submodules.
After removing this submodule, the context-aware represenations will be directly fed into the gated mechanism,
which is unreasonable.

\subsection{Results}
\begin{table*}[t]
  \begin{center}
    \resizebox{\textwidth}{!}{
      \begin{tabular}{|c|c|c|c|c|c|c|c|c|c|c|c|c|}
      \hline
      \multirow{2}{*}{\textbf{Models}} & \multicolumn{4}{c|}{\textbf{E-Commerce Corpus}} & \multicolumn{4}{c|}{\textbf{Douban Corpus}}   & \multicolumn{4}{c|}{\textbf{Zh50w Corpus}}    \\ \cline{2-13} 
                                       & \textbf{$R_{10}@1$}  & \textbf{$R_{10}@2$}  & \textbf{$R_{10}@5$} & \textbf{MRR} & \textbf{$R_{10}@1$} & \textbf{$R_{10}@2$} & \textbf{$R_{10}@5$} & \textbf{MRR} & \textbf{$R_{10}@1$} & \textbf{$R_{10}@2$} & \textbf{$R_{10}@5$} & \textbf{MRR} \\ \hline
      \textbf{bi-encoder}              & 0.71       & 0.865      & \textbf{0.972}     & 0.8221    &  0.3373         & 0.5457          & 0.8576          & 0.5477          & 0.2335          & 0.3429          & 0.5647          & 0.401           \\ \hline
      \textbf{bi-encoder+SCM}          & \textbf{0.744}      & \textbf{0.868}      & 0.97      & \textbf{0.8394}    &  \textbf{0.3643}         & \textbf{0.5592}          & \textbf{0.8762}          & \textbf{0.5628}          & \textbf{0.2408}          & \textbf{0.3526}          & \textbf{0.5794}          & \textbf{0.4096}          \\ \hline \hline
      \textbf{poly-encoder}            & 0.718      & 0.871      & 0.971     & 0.8275    &  0.3418         & 0.5487          & 0.8621          & 0.55            & 0.2308          & 0.3519          & 0.573           & 0.4025          \\ \hline
      \textbf{poly-encoder+SCM}        & \textbf{0.794}      & \textbf{0.877}      & \textbf{0.973}     & \textbf{0.8447}    &  \textbf{0.3583}         & \textbf{0.5637}     & \textbf{0.8636}          & \textbf{0.5627}          & \textbf{0.2505}          & \textbf{0.3532}          & \textbf{0.5851}          & \textbf{0.4152}          \\ \hline
      \end{tabular}
    }
    \caption{The overall comparison on E-Commerce, Douban, and Zh50w Corpus test datasets (10 candidate responses).}
    \label{tab:3}
  \end{center}
\end{table*}

\subsubsection{Overall Comparison}
Table \ref{tab:3} shows the overall comparison among all the compared models on three popular
response selection corpus.
Referring to the results in Table \ref{tab:3}, 
our proposed SCM module significantly boosts the performance 
of these two state-of-the-art retrieval-based open-domain dialog models,
which proves the effectiveness of SCM module.
It can be found that \textbf{bi-encoder+SCM} and \textbf{poly-encoder+SCM} 
achieve more than 3\% average absolute improvement on $R_{10}@1$ 
compared with with \textbf{bi-encoder} and \textbf{poly-encoder} models,
which demonstrate that our proposed SCM module could effectively boost the performance.

\subsubsection{Extended Experiments}
In E-Commerce, Douban, Zh50w Corpus, there are only 10 candidates (1 positive and 9 negative) for a given conversation context.
But in real scenarios, hundreds candidate responses are very usual.
Intuitively, more candidate responses could provide lots of nosies,
and a large number of candidate responses information will easily 
drown out valuable information,
which may bring great difficulties to our proposed SCM module.
So whether our proposed model is still better than the base models on real scenarios still remains questionable.
In order to answer this question, in this subsection,
we extend more candidate responses for one given conversation context to make the test dataset more difficult,
which is very close to the real scenarios.
Specifically, first of all, we save all the utterances in each corpus into the pre-constructed database.
Then, the mature toolkit Elasticsearch\footnote{\url{https://github.com/elastic/elasticsearch}} 
is used to retrieve lots of coherent utterances from the pre-constructed database,
which will be treated as the extended negative samples.

In this paper, we separately test 50, 100, 150, 200, 250, 300 candidate responses for one given conversation context.
As shown in Table \ref{tab:4}, we can make the following conclusions:
\begin{itemize}
  \item Though there are more candidate responses for one given conversation context, 
  our proposed SCM module still boosts the performance further in most cases.
  Even when the number of the candidate responses is extreme 300, 
  our proposed model still outperforms the corresponding base model,
  which proves the robustness of SCM module.
  \item It can be found that, in some cases, 
  our proposed model is slightly worse than its corresponding base model
  on $R_{x}@1$ and $R_{x}@2$ metric.
  After serious analysis, we think the reason is that, 
  the elasticsearch toolkit retrieves some very good candidate responses from the pre-constructed database
  as the extended samples. 
  Referring to the results on $R_{x}@5$, $R_{x}@10$, and MRR metric,
  even there are some good responses in the candidate set,
  our proposed model can still improve ranking of the ground-truth significantly,
  which is very important in real scene.
\end{itemize}

\begin{table*}[h]
  \begin{center}
    \resizebox{\textwidth}{!}{
      \begin{tabular}{|c|c|c|c|c|c|c|c|c|c|c|c|c|c|c|c|}
      \hline
      \multirow{2}{*}{\textbf{50 Candidate response}} & \multicolumn{5}{c|}{\textbf{E-Commerce Corpus}} & \multicolumn{5}{c|}{\textbf{Douban Corpus}}   & \multicolumn{5}{c|}{\textbf{Zh50w Corpus}}    \\ \cline{2-16} 
                                       & \textbf{$R_{50}@1$}  & \textbf{$R_{50}@2$}  & \textbf{$R_{50}@5$} & \textbf{$R_{50}@10$} & \textbf{MRR} & \textbf{$R_{50}@1$} & \textbf{$R_{50}@2$} & \textbf{$R_{50}@5$} & \textbf{$R_{50}@10$} & \textbf{MRR} & \textbf{$R_{50}@1$} & \textbf{$R_{50}@2$} & \textbf{$R_{50}@5$} & \textbf{$R_{50}@10$} & \textbf{MRR} \\ \hline
      \textbf{bi-encoder}              & 0.484              & 0.624             & 0.747             & 0.842                & 0.6088              & 0.0915          & \textbf{0.1499} & 0.2414          & 0.3193          & 0.182           & 0.0781          & 0.1348           & 0.2138          & 0.3052          & 0.1643          \\ \hline
      \textbf{bi-encoder+SCM}          & \textbf{0.505}     & \textbf{0.65}     & \textbf{0.761}    & \textbf{0.844}       & \textbf{0.627}      & \textbf{0.0915} & 0.1469          & \textbf{0.2489} & \textbf{0.3508} & \textbf{0.1851} & \textbf{0.0824} & \textbf{0.1394}  & \textbf{0.2275} & \textbf{0.3172} & \textbf{0.1715} \\ \hline \hline
      \textbf{poly-encoder}            & 0.505              & 0.621             & 0.744             & 0.835                & 0.6176              & 0.0945          & \textbf{0.1544} & 0.2384          & 0.3343          & 0.1842          & 0.0754          & 0.1328           & 0.2171          & 0.3099          & 0.1636          \\ \hline
      \textbf{poly-encoder+SCM}        & \textbf{0.54}      & \textbf{0.66}     & \textbf{0.789}    & \textbf{0.861}       & \textbf{0.6522}     & \textbf{0.1034} & 0.1499          & \textbf{0.2534} & \textbf{0.3613} & \textbf{0.1929} & \textbf{0.0801} & \textbf{0.1381}  & \textbf{0.2285} & \textbf{0.3225} & \textbf{0.1701} \\ \hline
      \end{tabular}
    }

    \resizebox{\textwidth}{!}{
      \begin{tabular}{|c|c|c|c|c|c|c|c|c|c|c|c|c|c|c|c|}
      \hline
      \multirow{2}{*}{\textbf{100 Candidate response}} & \multicolumn{5}{c|}{\textbf{E-Commerce Corpus}} & \multicolumn{5}{c|}{\textbf{Douban Corpus}}   & \multicolumn{5}{c|}{\textbf{Zh50w Corpus}}    \\ \cline{2-16} 
                                       & \textbf{$R_{100}@1$}  & \textbf{$R_{100}@2$}  & \textbf{$R_{100}@5$} & \textbf{$R_{100}@10$} & \textbf{MRR} & \textbf{$R_{100}@1$} & \textbf{$R_{100}@2$} & \textbf{$R_{100}@5$} & \textbf{$R_{100}@10$} & \textbf{MRR} & \textbf{$R_{100}@1$} & \textbf{$R_{100}@2$} & \textbf{$R_{100}@5$} & \textbf{$R_{100}@10$} & \textbf{MRR} \\ \hline
      \textbf{bi-encoder}              & 0.45              & 0.555             & 0.692             & 0.77                 & 0.5609              &  \textbf{0.078}& \textbf{0.1214} & 0.1889          &  0.2519         & \textbf{0.1453} & 0.0627          & 0.1057             & 0.1681          & 0.2382          & 0.1287       \\ \hline
      \textbf{bi-encoder+SCM}          & \textbf{0.456}    & \textbf{0.576}    & \textbf{0.701}    & \textbf{0.781}       & \textbf{0.5706}     &  0.072         &  0.1154         & \textbf{0.1904} & \textbf{0.2729} & 0.1439          & \textbf{0.064}  & \textbf{0.1091}    & \textbf{0.1808} & \textbf{0.2445} & \textbf{0.1331}       \\ \hline \hline
      \textbf{poly-encoder}            & 0.458             & 0.571             & 0.681             & 0.766                & 0.5674              &  0.072         & \textbf{0.1199} & 0.1904          &  0.2609         & 0.1426          & 0.0594          & 0.1031             & 0.1738          & 0.2358          & 0.1274       \\ \hline
      \textbf{poly-encoder+SCM}        & \textbf{0.493}    & \textbf{0.594}    & \textbf{0.715}    & \textbf{0.804}       & \textbf{0.598}      & \textbf{0.0855}&  0.1184         & \textbf{0.1934} & \textbf{0.2789} & \textbf{0.1518} & \textbf{0.0624} & \textbf{0.1101}    & \textbf{0.1831} & \textbf{0.2488} & \textbf{0.133}        \\ \hline
      \end{tabular}
    }

    \resizebox{\textwidth}{!}{
      \begin{tabular}{|c|c|c|c|c|c|c|c|c|c|c|c|c|c|c|c|}
      \hline
      \multirow{2}{*}{\textbf{150 Candidate response}} & \multicolumn{5}{c|}{\textbf{E-Commerce Corpus}} & \multicolumn{5}{c|}{\textbf{Douban Corpus}}   & \multicolumn{5}{c|}{\textbf{Zh50w Corpus}}    \\ \cline{2-16} 
                                       & \textbf{$R_{150}@1$}  & \textbf{$R_{150}@2$}  & \textbf{$R_{150}@5$} & \textbf{$R_{150}@10$} & \textbf{MRR} & \textbf{$R_{150}@1$} & \textbf{$R_{150}@2$} & \textbf{$R_{150}@5$} & \textbf{$R_{150}@10$} & \textbf{MRR} & \textbf{$R_{150}@1$} & \textbf{$R_{150}@2$} & \textbf{$R_{150}@5$} & \textbf{$R_{150}@10$} & \textbf{MRR} \\ \hline
      \textbf{bi-encoder}              & 0.426             & 0.533             & 0.654             & 0.732            & 0.5345          & \textbf{0.0765} & 0.1004          & 0.1559          & 0.2144          & 0.1267          & 0.0534            & 0.0921            & 0.1498          & 0.2045          & 0.1118               \\ \hline
      \textbf{bi-encoder+SCM}          & \textbf{0.436}    & \textbf{0.557}    & \textbf{0.676}    & \textbf{0.739}   & \textbf{0.5491} & 0.0675          & \textbf{0.1019} & \textbf{0.1709} & \textbf{0.2249} & \textbf{0.1272} & \textbf{0.0544}   & \textbf{0.0967}   & \textbf{0.1621} & \textbf{0.2195} & \textbf{0.1164}      \\ \hline \hline
      \textbf{poly-encoder}            & 0.433             & 0.541             & 0.654             & 0.724            & 0.5381          & 0.063           & 0.0975          & 0.1529          & 0.2174          & 0.12            & 0.0534            & 0.0937            & 0.1544          & 0.2095          & 0.113                \\ \hline
      \textbf{poly-encoder+SCM}        & \textbf{0.471}    & \textbf{0.579}    & \textbf{0.682}    & \textbf{0.761}   & \textbf{0.5734} & \textbf{0.0765} & \textbf{0.1079} & \textbf{0.1784} & \textbf{0.2249} & \textbf{0.1324} & \textbf{0.0567}   & \textbf{0.0941}   & \textbf{0.1558} & \textbf{0.2211} & \textbf{0.116}       \\ \hline
      \end{tabular}
    }

    \resizebox{\textwidth}{!}{
      \begin{tabular}{|c|c|c|c|c|c|c|c|c|c|c|c|c|c|c|c|}
      \hline
      \multirow{2}{*}{\textbf{200 Candidate response}} & \multicolumn{5}{c|}{\textbf{E-Commerce Corpus}} & \multicolumn{5}{c|}{\textbf{Douban Corpus}}   & \multicolumn{5}{c|}{\textbf{Zh50w Corpus}}    \\ \cline{2-16} 
                                       & \textbf{$R_{200}@1$}  & \textbf{$R_{200}@2$}  & \textbf{$R_{200}@5$} & \textbf{$R_{200}@10$} & \textbf{MRR} & \textbf{$R_{200}@1$} & \textbf{$R_{200}@2$} & \textbf{$R_{200}@5$} & \textbf{$R_{200}@10$} & \textbf{MRR} & \textbf{$R_{200}@1$} & \textbf{$R_{200}@2$} & \textbf{$R_{200}@5$} & \textbf{$R_{200}@10$} & \textbf{MRR} \\ \hline
      \textbf{bi-encoder}              & 0.405             & 0.508             & 0.637             & 0.721            & 0.513            & \textbf{0.072}  & \textbf{0.1034} & 0.1484          & 0.2024          & 0.1212         & 0.049             & 0.0827          & 0.1371          & 0.1821          & 0.101         \\ \hline
      \textbf{bi-encoder+SCM}          & \textbf{0.405}    & \textbf{0.537}    & \textbf{0.651}    & \textbf{0.733}   & \textbf{0.5226}  & 0.0645          & 0.1019          & \textbf{0.1634} & \textbf{0.2084} & \textbf{0.122} & \textbf{0.05}     & \textbf{0.0887} & \textbf{0.1448} & \textbf{0.2008} & \textbf{0.1056}        \\ \hline \hline
      \textbf{poly-encoder}            & 0.422             & 0.515             & 0.642             & 0.709            & 0.5229           & 0.0675          & 0.1034          & 0.1484          & 0.2024          & 0.119          & 0.0464            & 0.0801          & 0.1384          & 0.1918          & 0.1008        \\ \hline
      \textbf{poly-encoder+SCM}        & \textbf{0.444}    & \textbf{0.553}    & \textbf{0.676}    & \textbf{0.738}   & \textbf{0.5499}  & \textbf{0.0705} & \textbf{0.1064} & \textbf{0.1589} & \textbf{0.2129} & \textbf{0.1249}& \textbf{0.0477}   & \textbf{0.0837} & \textbf{0.1474} & \textbf{0.1978} & \textbf{0.1041}        \\ \hline
      \end{tabular}
    }

    \resizebox{\textwidth}{!}{
      \begin{tabular}{|c|c|c|c|c|c|c|c|c|c|c|c|c|c|c|c|}
      \hline
      \multirow{2}{*}{\textbf{250 Candidate response}} & \multicolumn{5}{c|}{\textbf{E-Commerce Corpus}} & \multicolumn{5}{c|}{\textbf{Douban Corpus}}   & \multicolumn{5}{c|}{\textbf{Zh50w Corpus}}    \\ \cline{2-16} 
                                       & \textbf{$R_{250}@1$}  & \textbf{$R_{250}@2$}  & \textbf{$R_{250}@5$} & \textbf{$R_{250}@10$} & \textbf{MRR} & \textbf{$R_{250}@1$} & \textbf{$R_{250}@2$} & \textbf{$R_{250}@5$} & \textbf{$R_{250}@10$} & \textbf{MRR} & \textbf{$R_{250}@1$} & \textbf{$R_{250}@2$} & \textbf{$R_{250}@5$} & \textbf{$R_{250}@10$} & \textbf{MRR} \\ \hline
      \textbf{bi-encoder}              & \textbf{0.405}   & \textbf{0.515}    & 0.613             & 0.702            & 0.5076          & 0.06           & \textbf{0.0975} & 0.1394          & 0.1784          & 0.109           & 0.0454           & 0.0767            & 0.1281          & 0.1738          & 0.0945      \\ \hline
      \textbf{bi-encoder+SCM}          & 0.401            & 0.509             & \textbf{0.629}    & \textbf{0.709}   & \textbf{0.51}   & \textbf{0.069} & 0.0915          & \textbf{0.1514} & \textbf{0.1964} & \textbf{0.1158} & \textbf{0.0464}  & \textbf{0.0801}   & \textbf{0.1382} & \textbf{0.1841} & \textbf{0.0976}      \\ \hline \hline
      \textbf{poly-encoder}            & 0.411            & 0.511             & 0.618             & 0.686            & 0.5105          & 0.054          & 0.0945          & 0.1428          & 0.1814          & 0.1057          & \textbf{0.0457}  & 0.0767            & 0.1331          & 0.1791          & 0.0955       \\ \hline
      \textbf{poly-encoder+SCM}        & \textbf{0.435}   & \textbf{0.546}    & \textbf{0.645}    & \textbf{0.715}   & \textbf{0.5377} & \textbf{0.066} & \textbf{0.1049} & \textbf{0.1484} & \textbf{0.1934} & \textbf{0.1179} & 0.0444           & \textbf{0.0787}   & \textbf{0.1344} & \textbf{0.1868} & \textbf{0.0962}       \\ \hline
      \end{tabular}
    }

    \resizebox{\textwidth}{!}{
      \begin{tabular}{|c|c|c|c|c|c|c|c|c|c|c|c|c|c|c|c|}
      \hline
      \multirow{2}{*}{\textbf{300 Candidate response}} & \multicolumn{5}{c|}{\textbf{E-Commerce Corpus}} & \multicolumn{5}{c|}{\textbf{Douban Corpus}}   & \multicolumn{5}{c|}{\textbf{Zh50w Corpus}}    \\ \cline{2-16} 
                                       & \textbf{$R_{300}@1$}  & \textbf{$R_{300}@2$}  & \textbf{$R_{300}@5$} & \textbf{$R_{300}@10$} & \textbf{MRR} & \textbf{$R_{300}@1$} & \textbf{$R_{300}@2$} & \textbf{$R_{300}@5$} & \textbf{$R_{300}@10$} & \textbf{MRR} & \textbf{$R_{300}@1$} & \textbf{$R_{300}@2$} & \textbf{$R_{300}@5$} & \textbf{$R_{300}@10$} & \textbf{MRR} \\ \hline
      \textbf{bi-encoder}              & 0.38             & \textbf{0.484}    & 0.588             & 0.681            & 0.4834          & 0.0525          & \textbf{0.099} & 0.1334          & 0.1649          & 0.102          & 0.0424           & 0.0757          & 0.1234          & 0.1598          & 0.0895         \\ \hline
      \textbf{bi-encoder+SCM}          & \textbf{0.383}   & 0.482             & \textbf{0.608}    & \textbf{0.694}   & \textbf{0.4897} & \textbf{0.0585} & 0.0915         & \textbf{0.1424} & \textbf{0.1844} & \textbf{0.107} & \textbf{0.0434}  & \textbf{0.0777} & \textbf{0.1261} & \textbf{0.1755} & \textbf{0.0923}         \\ \hline \hline
      \textbf{poly-encoder}            & 0.399            & 0.483             & 0.6               & 0.677            & 0.494           & 0.051           & 0.0945         & 0.1289          & 0.1769          & 0.1001         & 0.0404           & \textbf{0.073}  & 0.1264          & 0.1698          & 0.0885         \\ \hline
      \textbf{poly-encoder+SCM}        & \textbf{0.414}   & \textbf{0.514}    & \textbf{0.628}    & \textbf{0.698}   & \textbf{0.5171} & \textbf{0.06}   & \textbf{0.0945}& \textbf{0.1439} & \textbf{0.1904} & \textbf{0.1098}& \textbf{0.042}   & 0.0717          & \textbf{0.1278} & \textbf{0.1761} & \textbf{0.0902}         \\ \hline
      \end{tabular}
    }
    \caption{The extended experiment on E-Commerce, Douban, and Zh50w Corpus test datasets (50, 100, 150, 200, 250, 300 candidate responses).}
    \label{tab:4}
  \end{center}
\end{table*}

\begin{table}[h]
  \begin{center}
  \resizebox{0.5\textwidth}{!}{
        \begin{tabular}{c|c|c|c|c}
        \hline
        \textbf{Model} & \textbf{$R_{10}@1$} & \textbf{$R_{10}@2$} & \textbf{$R_{10}@5$} & \textbf{MRR} \\ \hline
        \textbf{poly-encoder+SCM}       & 0.766               & 0.882               & 0.975                & 0.8537    \\ \hline
        \textbf{-\{context-aware\}}     &  0.708   &  0.854    &    0.968  &   0.8178  \\ 
        \textbf{-gated}     &  0.73   &  0.859    &  0.97    &  0.8298   \\ \hline
        \textbf{poly-encoder}    & 0.71    & 0.865     & 0.972     & 0.8221    \\ \hline
        \end{tabular}
  }
  \caption{Ablation study of \textbf{context-aware represenation} and \textbf{gated mechanism} submodules in proposed SCM module on E-Commerce Corpus.}
  \label{tab:5}
  \end{center}
\end{table}

\subsubsection{Adversarial Experiment}

Previous work \cite{Whang2020DoRS} shows that,
BERT-based models tends to give higher
probability score to the response which is more semantically
related to the context rather than consistent response.
Following the previous work, we conduct the adversarial experiment
to investigate whether these models are trained properly.
Specifically, for one test sample that contains 1 conversation context,
1 ground-truth response, and 9 negative responses,
we replace one of the negative responses with an adversarial response that is randomly extracted from the conversation context,
and we could obtain 1 ground-truth response, 1 adversarial response, 8 negative responses.
It should be noted that the adversarial response is semantically related to the context,
but is not appropriate.
Then, these models are measured on the these new test samples.
Referring to the Table, we can make the following conclusions:
\begin{itemize}
  \item The performance of all of the models drop sharply, which demonstrates that
  all of the BERT-based models confused by the adversarial responses.
  Researchers should pay more attention to this fatal weakness of current state-of-the-art models.
  \item It can be found that, 
  our proposed models still significantly outperforms these base models.
  For example,
  The results prove the robustness of our proposed SCM module.
\end{itemize}

\begin{table}[h]
  \resizebox{0.5\textwidth}{!}{
    \subtable[Adversarial experiment on E-Commerce Corpus.]{
      \begin{tabular}{c|c|c|c|c}
      \hline
      \textbf{models} & \textbf{$R_{10}@1$} & \textbf{$R_{10}@2$} & \textbf{$R_{10}@5$} & \textbf{MRR} \\ \hline
        poly-encoder    &  0.601              &    0.83             &  0.968              &  0.7588         \\ \hline
        poly-encoder+SCM&                     &                     &                     &                 \\ \hline \hline
        bi-encoder      &  0.58               &    0.821            &  0.968              &  0.7466         \\ \hline
        bi-encoder+SCM  &  0.601              &    0.84             &  0.963              &  0.7605         \\ \hline
      \end{tabular}
    }
  }
  \qquad
  \resizebox{0.5\textwidth}{!}{
    \subtable[Adversarial experiment on Douban Corpus.]{
      \begin{tabular}{c|c|c|c|c}
      \hline
      \textbf{models} & \textbf{$R_{10}@1$} & \textbf{$R_{10}@2$} & \textbf{$R_{10}@5$} & \textbf{MRR} \\ \hline
        poly-encoder    &  0.1514             &    0.4123           &  0.7976             &  0.4175         \\ \hline
        poly-encoder+SCM&  0.1649             &    0.4513           &  0.8276             &  0.435               \\ \hline \hline
        bi-encoder      &  0.58               &    0.821            &  0.968              &  0.7466         \\ \hline
        bi-encoder+SCM  &  0.601              &    0.84             &  0.963              &  0.7605         \\ \hline
      \end{tabular}
    }
  }
  \caption{}
  \label{}
\end{table}

\subsubsection{Ablation Study}
In this subsection, we conduct the ablation study to analyze the contributions of two submodules in our proposed SCM module:
(1) context-aware represenation; (2) gated mechanism.
Due to the page limitation, we only show partial results in Table, and more results can be found in \textit{Appendix}.
It should be noted that the results in three datasets are consistent.
Referring to the Table, we can make the following conclusions:
\begin{itemize}
  \item \textbf{Context-aware represenation} and \textbf{gated mechanism} are necessary and important, 
  and the performance drops sharply when these submodules are removed.
  \item Based on the size of the performance decline, 
  it can be observed that, 
  the most important submodule in our proposed SCM module is 
  \textbf{context-aware represenation} and then \textbf{gated mechanism}.
\end{itemize} 

\subsubsection{Hyperparameters Analysis}


In this subsection, 
we will analyze three important hyperparameters of our proposed SCM module in \textbf{poly-encoder+SCM} model on E-Commerce Corpus: 
(1) the number of the transformer layers $n$;
(2) the number of the heads in each transformer layer $n_{head}$;
(3) the dimension feed-forward in each transformer layer $dim_{ffd}$.
Due to the page limitation, we only show partial results in Table, more details can be found in \textit{Appendix}.
Referring to the results in Table \ref{tab:5}, we can make the following conclusions:
\begin{itemize}
  \item For different datasets, the optimial parameters setting for our proposed SCM module is a little bit different.
  But, except for few cases, it can be found that our proposed SCM boosts the performance further in most of parameters settings,
  which prove the robustness of our model.
  \item For these three hyperparameters, too 
\end{itemize}

\begin{table}[h]
  \centering
  \resizebox{0.5\textwidth}{!}{
    \subtable[Hyperparameter $n$. $n_{head}=8$ and $dim_{ffd}=512$.]{
        \begin{tabular}{|c|c|c|c|c|}
        \hline
        \textbf{poly-encoder+SCM} & \textbf{$R_{10}@1$} & \textbf{$R_{10}@2$} & \textbf{$R_{10}@5$} & \textbf{MRR} \\ \hline
        \textbf{poly-encoder}     &    0.718  & 0.871     & 0.971     & 0.8275    \\ \hline
        \textbf{$n=2$}            &    0.738  & 0.878     & 0.973     & 0.8392    \\ \hline
        \textbf{$n=4$}            &    \textbf{0.794}  & 0.877     & \textbf{0.973}     & 0.8305    \\ \hline
        \textbf{$n=6$}            &    0.703  & 0.828     & 0.962     & 0.8088    \\ \hline
        \textbf{$n=8$}            &    0.752  & \textbf{0.885}     & 0.967     & \textbf{0.8468}    \\ \hline
        \end{tabular}
    }
  }
  \qquad
  \resizebox{0.5\textwidth}{!}{
    \subtable[Hyperparameter $n_{head}$. $n=4$ and $dim_{ffd}=512$.]{
      \begin{tabular}{|c|c|c|c|c|}
        \hline
        \textbf{poly-encoder+SCM} & \textbf{$R_{10}@1$} & \textbf{$R_{10}@2$} & \textbf{$R_{10}@5$} & \textbf{MRR} \\ \hline
        \textbf{poly-encoder}     & 0.718     & 0.871     & 0.971     & 0.8275    \\ \hline
        \textbf{$n_{head}=2$}     & 0.727     & 0.868     & 0.967     & 0.830     \\ \hline
        \textbf{$n_{head}=4$}     & 0.729     & 0.859     & 0.971     & 0.8304    \\ \hline
        \textbf{$n_{head}=6$}     & 0.748     & \textbf{0.891}     & 0.97      & \textbf{0.8459}    \\ \hline
        \textbf{$n_{head}=8$}     & \textbf{0.794}     & 0.877     & \textbf{0.973}     & 0.8305    \\ \hline
        \end{tabular}
    }
  }
  \qquad
  \resizebox{0.5\textwidth}{!}{
    \subtable[Hyperparameter $dim_{ffd}$. $n=4$ and $n_{head}=8$.]{
      \begin{tabular}{|c|c|c|c|c|}
        \hline
        \textbf{poly-encoder+SCM} & \textbf{$R_{10}@1$} & \textbf{$R_{10}@2$} & \textbf{$R_{10}@5$} & \textbf{MRR} \\ \hline
        \textbf{poly-encoder}       & 0.718     & 0.871     & 0.971     & 0.8275    \\ \hline
        \textbf{$dim_{ffd}=128$}    & 0.739     & 0.877     & 0.974     & 0.8389    \\ \hline
        \textbf{$dim_{ffd}=512$}    & \textbf{0.794}     & 0.877     & 0.973     & 0.8447    \\ \hline
        \textbf{$dim_{ffd}=1024$}   & 0.724     & 0.871     & 0.966     & 0.8299    \\ \hline
        \textbf{$dim_{ffd}=2048$}   & 0.766     & \textbf{0.882}     & \textbf{0.975}     & \textbf{0.8537}    \\ \hline
        \end{tabular}
    }
  }
  \caption{Three important hyperparameters analysis of \textbf{poly-encoder+SCM} model on E-Commerce Corpus.}
  \label{tab:5}
\end{table}

\section{Conclusions}

In this paper, we propose a novel and plug-in self-attention comparison module (SCM) for the state-of-the-art retrieval-based open-domain dialog systems.
Extensive experiemnts demonstrate that our proposed SCM module significantly boosts the performance on three popular open-domain response selection corpus.
Moreover, the extended experiments that we conduct also prove the effectiveness of SCM module in real scenarios.
Besides, we have publicly released the source codes of our proposed model in this paper for future research.

\bibliography{anthology,acl2020}
\bibliographystyle{acl_natbib}

\appendix

\section{Appendices}

\subsection{Ablation Study}
\begin{table}[h]
  \centering
  \resizebox{0.5\textwidth}{!}{
    \subtable[Ablation Study on Douban Study]{
        \begin{tabular}{c|c|c|c|c}
        \hline
        \textbf{Model} & \textbf{$R_{10}@1$} & \textbf{$R_{10}@2$} & \textbf{$R_{10}@5$} & \textbf{MRR} \\ \hline
        \textbf{poly-encoder+SCM}       & 0.766               & 0.882               & 0.975                & 0.8537    \\ \hline
        \textbf{-\{context-aware\}}     &  0.708   &  0.854    &    0.968  &   0.8178  \\ 
        \textbf{-gated}     &  0.73   &  0.859    &  0.97    &  0.8298   \\ \hline
        \textbf{poly-encoder}    & 0.71    & 0.865     & 0.972     & 0.8221    \\ \hline
        \end{tabular}
    }
  }
  \qquad
  \resizebox{0.5\textwidth}{!}{
    \subtable[Ablation Study on Zh50w Study]{
        \begin{tabular}{c|c|c|c|c}
        \hline
        \textbf{Model} & \textbf{$R_{10}@1$} & \textbf{$R_{10}@2$} & \textbf{$R_{10}@5$} & \textbf{MRR} \\ \hline
        \textbf{poly-encoder+SCM}       & 0.766               & 0.882               & 0.975                & 0.8537    \\ \hline
        \textbf{-\{context-aware\}}     &  0.708   &  0.854    &    0.968  &   0.8178  \\ 
        \textbf{-gated}     &  0.73   &  0.859    &  0.97    &  0.8298   \\ \hline
        \textbf{poly-encoder}    & 0.71    & 0.865     & 0.972     & 0.8221    \\ \hline
        \end{tabular}
    }
  }
  \caption{Ablation study of \textbf{context-aware represenation} and \textbf{gated mechanism} submodules in proposed SCM module. Base model is \textbf{poly-encoder}.}
  \label{tab:5}
\end{table}

\subsection{Hyperparameters}

\begin{table}[h]
  \centering
  \resizebox{0.5\textwidth}{!}{
    \subtable[Hyperparameter $n$. $n_{head}=8$ and $dim_{ffd}=512$.]{
        \begin{tabular}{|c|c|c|c|c|}
        \hline
        \textbf{poly-encoder+SCM} & \textbf{$R_{10}@1$} & \textbf{$R_{10}@2$} & \textbf{$R_{10}@5$} & \textbf{MRR} \\ \hline
        \textbf{poly-encoder}     &    0.3343&0.5262&0.8501&0.5372    \\ \hline
        \textbf{$n=2$}            &    0.3463&0.5502&0.8681&0.553   \\ \hline
        \textbf{$n=4$}            &    0.3718  & 0.5517     & 0.8591     & 0.5652    \\ \hline
        \textbf{$n=6$}            &    0.3523&0.5442&0.8756&0.5555    \\ \hline
        \textbf{$n=8$}            &    0.3178&0.4936&0.8276&0.5201 \\ \hline
        \end{tabular}
    }
  }
  \qquad
  \resizebox{0.5\textwidth}{!}{
    \subtable[Hyperparameter $n_{head}$. $n=4$ and $dim_{ffd}=512$.]{
      \begin{tabular}{|c|c|c|c|c|}
        \hline
        \textbf{poly-encoder+SCM} & \textbf{$R_{10}@1$} & \textbf{$R_{10}@2$} & \textbf{$R_{10}@5$} & \textbf{MRR} \\ \hline
        \textbf{poly-encoder}     & 0.3343&0.5262&0.8501&0.5372   \\ \hline
        \textbf{$n_{head}=2$}     & 0.3538&0.5502&0.8411&0.5545    \\ \hline
        \textbf{$n_{head}=4$}     & 0.3298&0.5427&0.8531&0.5428    \\ \hline
        \textbf{$n_{head}=6$}     & 0.3718     & 0.5517     & 0.8591      & 0.5652    \\ \hline
        \textbf{$n_{head}=8$}     & 0.3523&0.5472&0.8561&0.5548 \\ \hline
        \end{tabular}
    }
  }
  \qquad
  \resizebox{0.5\textwidth}{!}{
    \subtable[Hyperparameter $dim_{ffd}$. $n=4$ and $n_{head}=8$.]{
      \begin{tabular}{|c|c|c|c|c|}
        \hline
        \textbf{poly-encoder+SCM} & \textbf{$R_{10}@1$} & \textbf{$R_{10}@2$} & \textbf{$R_{10}@5$} & \textbf{MRR} \\ \hline
        \textbf{poly-encoder}       & 0.3343& 0.5262&0.8501&0.5372   \\ \hline
        \textbf{$dim_{ffd}=128$}    & 0.3508&0.5412&0.8546&0.5542    \\ \hline
        \textbf{$dim_{ffd}=512$}    & 0.3523&0.5472&0.8561&0.5548    \\ \hline
        \textbf{$dim_{ffd}=1024$}   & 0.3343&0.5112&0.8351&0.5352    \\ \hline
        \textbf{$dim_{ffd}=2048$}   & 0.3133&0.4813&0.8141&0.513 \\ \hline
        \end{tabular}
    }
  }
  \caption{Three important hyperparameters analysis of \textbf{poly-encoder+SCM} model on Douban Corpus.}
  \label{tab:5}
\end{table}

\begin{table}[h]
  \centering
  \resizebox{0.5\textwidth}{!}{
    \subtable[Hyperparameter $n$. $n_{head}=8$ and $dim_{ffd}=512$.]{
        \begin{tabular}{|c|c|c|c|c|}
        \hline
        \textbf{poly-encoder+SCM} & \textbf{$R_{10}@1$} & \textbf{$R_{10}@2$} & \textbf{$R_{10}@5$} & \textbf{MRR} \\ \hline
        \textbf{poly-encoder}     &    0.3343&0.5262&0.8501&0.5372    \\ \hline
        \textbf{$n=2$}            &    0.3463&0.5502&0.8681&0.553   \\ \hline
        \textbf{$n=4$}            &    0.3718  & 0.5517     & 0.8591     & 0.5652    \\ \hline
        \textbf{$n=6$}            &    0.3523&0.5442&0.8756&0.5555    \\ \hline
        \textbf{$n=8$}            &    0.3178&0.4936&0.8276&0.5201 \\ \hline
        \end{tabular}
    }
  }
  \qquad
  \resizebox{0.5\textwidth}{!}{
    \subtable[Hyperparameter $n_{head}$. $n=4$ and $dim_{ffd}=512$.]{
      \begin{tabular}{|c|c|c|c|c|}
        \hline
        \textbf{poly-encoder+SCM} & \textbf{$R_{10}@1$} & \textbf{$R_{10}@2$} & \textbf{$R_{10}@5$} & \textbf{MRR} \\ \hline
        \textbf{poly-encoder}     & 0.3343&0.5262&0.8501&0.5372   \\ \hline
        \textbf{$n_{head}=2$}     & 0.3538&0.5502&0.8411&0.5545    \\ \hline
        \textbf{$n_{head}=4$}     & 0.3298&0.5427&0.8531&0.5428    \\ \hline
        \textbf{$n_{head}=6$}     & 0.3718     & 0.5517     & 0.8591      & 0.5652    \\ \hline
        \textbf{$n_{head}=8$}     & 0.3523&0.5472&0.8561&0.5548 \\ \hline
        \end{tabular}
    }
  }
  \qquad
  \resizebox{0.5\textwidth}{!}{
    \subtable[Hyperparameter $dim_{ffd}$. $n=4$ and $n_{head}=8$.]{
      \begin{tabular}{|c|c|c|c|c|}
        \hline
        \textbf{poly-encoder+SCM} & \textbf{$R_{10}@1$} & \textbf{$R_{10}@2$} & \textbf{$R_{10}@5$} & \textbf{MRR} \\ \hline
        \textbf{poly-encoder}       & 0.3343& 0.5262&0.8501&0.5372   \\ \hline
        \textbf{$dim_{ffd}=128$}    & 0.3508&0.5412&0.8546&0.5542    \\ \hline
        \textbf{$dim_{ffd}=512$}    & 0.3523&0.5472&0.8561&0.5548    \\ \hline
        \textbf{$dim_{ffd}=1024$}   & 0.3343&0.5112&0.8351&0.5352    \\ \hline
        \textbf{$dim_{ffd}=2048$}   & 0.3133&0.4813&0.8141&0.513 \\ \hline
        \end{tabular}
    }
  }
  \caption{Three important hyperparameters analysis of \textbf{poly-encoder+SCM} model on Zh50w Corpus.}
  \label{tab:5}
\end{table}

\end{document}